\documentclass[sn-mathphys]{sn-jnl}

\graphicspath{ {./figures/} }
\usepackage{hyperref}
\usepackage{float}
\usepackage{verbatim} 

\restylefloat{figure}
\restylefloat{table}
\usepackage[utf8x]{inputenc}
\usepackage{caption}
\usepackage{fonts-tlwg}
\usepackage{mathtools}
\usepackage{amsmath}

\jyear{2021}%

\theoremstyle{thmstyleone}%
\theoremstyle{thmstyletwo}%

\theoremstyle{thmstylethree}%

\begin{document}

\title[TFS Recognition: Investigating MPH]{Thai Finger Spelling Recognition: Investigating MediaPipe Hands Potentials}


\author[1]{\fnm{Jinnavat} \sur{Sanalohit}}\email{jinnawat8@gmail.com}

\author[1]{\fnm{Tatpong} \sur{Katanyukul}}\email{tatpong@kku.ac.th}

\affil[1]{\orgdiv{Computer Engineering}, \orgname{Khon Kaen University}, \orgaddress{\street{Mitraparb Road}, \city{Khon Kaen}, \postcode{40002},  \country{Thailand}}}


\abstract{

Thai Finger Spelling (TFS) sign recognition could benefit a community of hearing-difficulty people in bridging to a major hearing population.
With a relatively large number of alphabets, TFS employs multiple signing schemes.
Two schemes of more common signing---static and dynamic single-hand signing, widely used in other sign languages---have been addressed in several previous works.
To complete the TFS sign recognition,
the remaining two of quite distinct signing schemes---static and dynamic point-on-hand signing---need to be sufficiently addressed.

With the advent of many off-the-shelf hand skeleton prediction models and
that training a model to recognize a sign language from scratch is expensive,
we explore an approach building upon 
recently launched
MediaPipe Hands (MPH).
MPH is a high-precision well-trained model for hand-keypoint detection. 
	We have investigated MPH on three TFS schemes: static-single-hand (S1), simplified dynamic-single-hand (S2) and static-point-on-hand (P1) schemes.
	Our results show that MPH can satisfactorily address single-hand schemes with accuracy of 84.57\% on both S1 and S2. 
	However, our finding reveals a shortcoming of MPH in addressing a point-on-hand scheme, whose accuracy is 23.66\% on P1
conferring to 69.19\% obtained from conventional classification trained from scratch.
This shortcoming has been investigated and attributed to self occlusion and handedness.
}




\maketitle


	\section{Introduction}
	Face-to-face communication is an important channel that can convey message, feeling, and personal connection. 
People with hearing impairments use two types of communication, which are reading-writing and a sign language. 
Reading and writing are slower and more formal than using a sign language. 
In addition, Antia et al.\cite{eni026} report that a hearing impaired person is often slower on reading and writing than hearing majority. 
A sign language remains relevant and essential for a hearing-impaired person in our current digital age as a spoken language for a hearing majority. 

According to the statistics of hearing impaired persons in 2018%
\cite{n_dep}, there are about 400,000 people, in Thailand, with hearing disabilities. 
While a number of Thai Sign Language (TSL) interpreters
is about 654 people.%
\footnote{This is estimated from a number of interpreters registered to the Department of Empowerment of Persons with Disabilities, Thailand\cite{n_interpreter}.} 
That is roughly one interpreter per 611 people with hearing disabilities.
Training sign language interpreters
takes considerable time and effort.
%
To mitigate the issue, an automatic translation system 
%
can provide supplementary service
to
bridge the communication gap between hearing impairments and the hearing majority.

A sign language is not a universally common language, e.g.,
the U.S. has American Sign Language (ASL); 
the U.K. has British Sign Language (BSL);
China has Chinese Sign Language (CSL); Thailand has TSL.
A sign language employs two signing modes: semantic and spelling signings.
A semantic signing uses hand and body gestures along with facial expressions to act out thoughts, meanings, attitudes, and feelings. 
A spelling signing uses finger and hand gestures to spell out a proper name or a word whose meaning has not yet been established in a particular language.
	
\begin{figure}[H]
		\centering
		\includegraphics[width=1\textwidth]{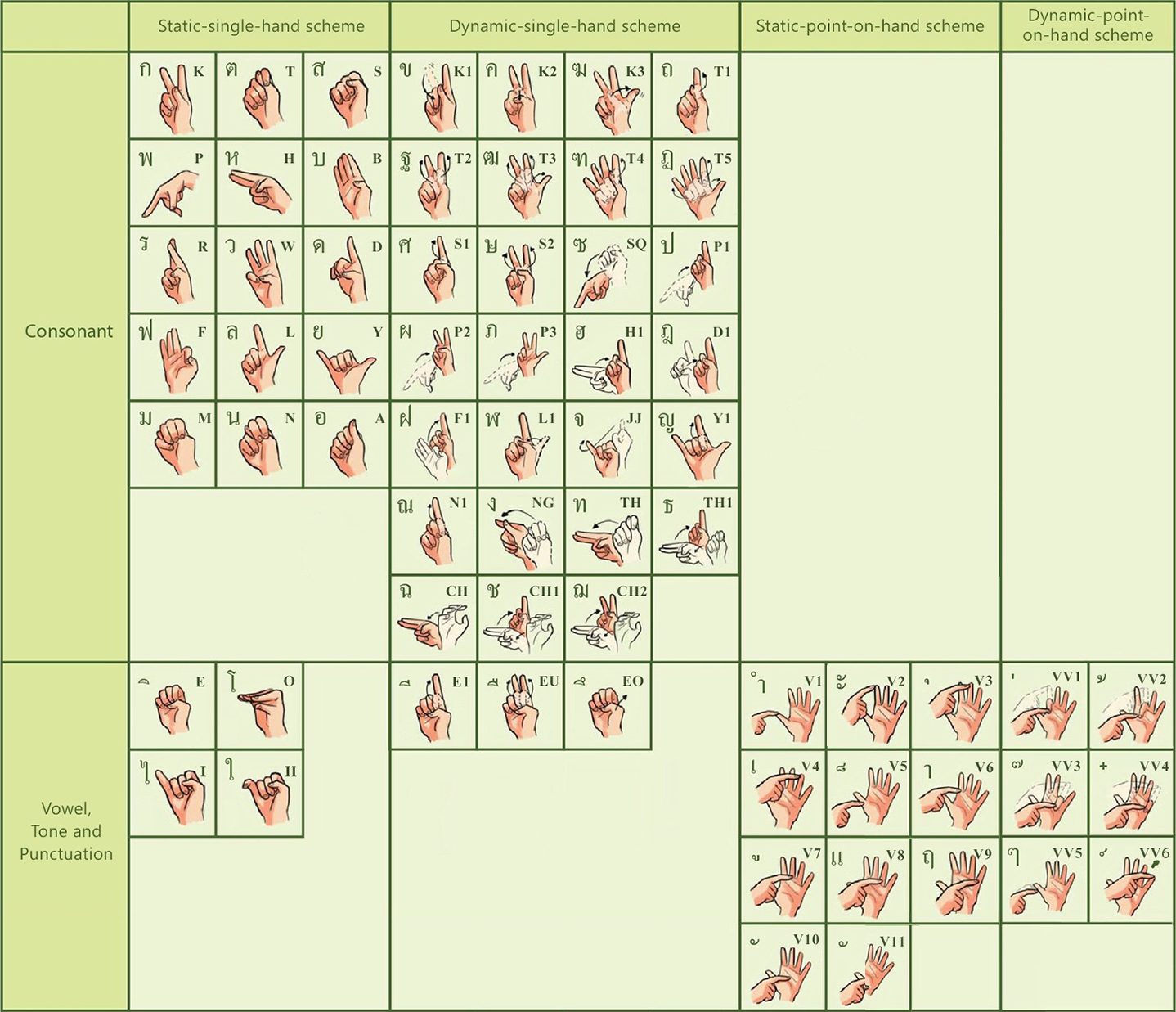}
		\caption{All TFS signs organized in four different schemes. Thai character is shown on the top-left of each cell, 
while an index symbol for easy reference is shown on the top-right.
TFS signs have been heavily influenced by American Finger Spelling (AFS) with many unique additions, including point-on-hand signing.}
		\label{fig:gesture-vs-scheme}
	\end{figure}

	Thai language has 42 consonants, 32 vowels and 4 intonation marks. 
To cope with a large set of Thai characters, Thai Finger Spell (TFS)---the spelling approach of TSL---uses various signing schemes. 
We categorize TFS, from pattern-recognition perspective, into four signing schemes:
static-single-hand, 
dynamic-single-hand, 
static-point-on-hand,
and dynamic-point-on-hand schemes. 
Figure~\ref{fig:gesture-vs-scheme} shows all TFS signings%
\footnote{%
The E and E1 signs are based on modern practice of TFS. 
A signer of an older convention may use a different practice, i.e.,  signing AFS E (E signing used in American Finger Spelling) following by AFS X to represent TFS E 
and signing AFS E to represent TFS E1.}.

\begin{figure}[H]
		\centering
		\includegraphics[width=0.5\textwidth]{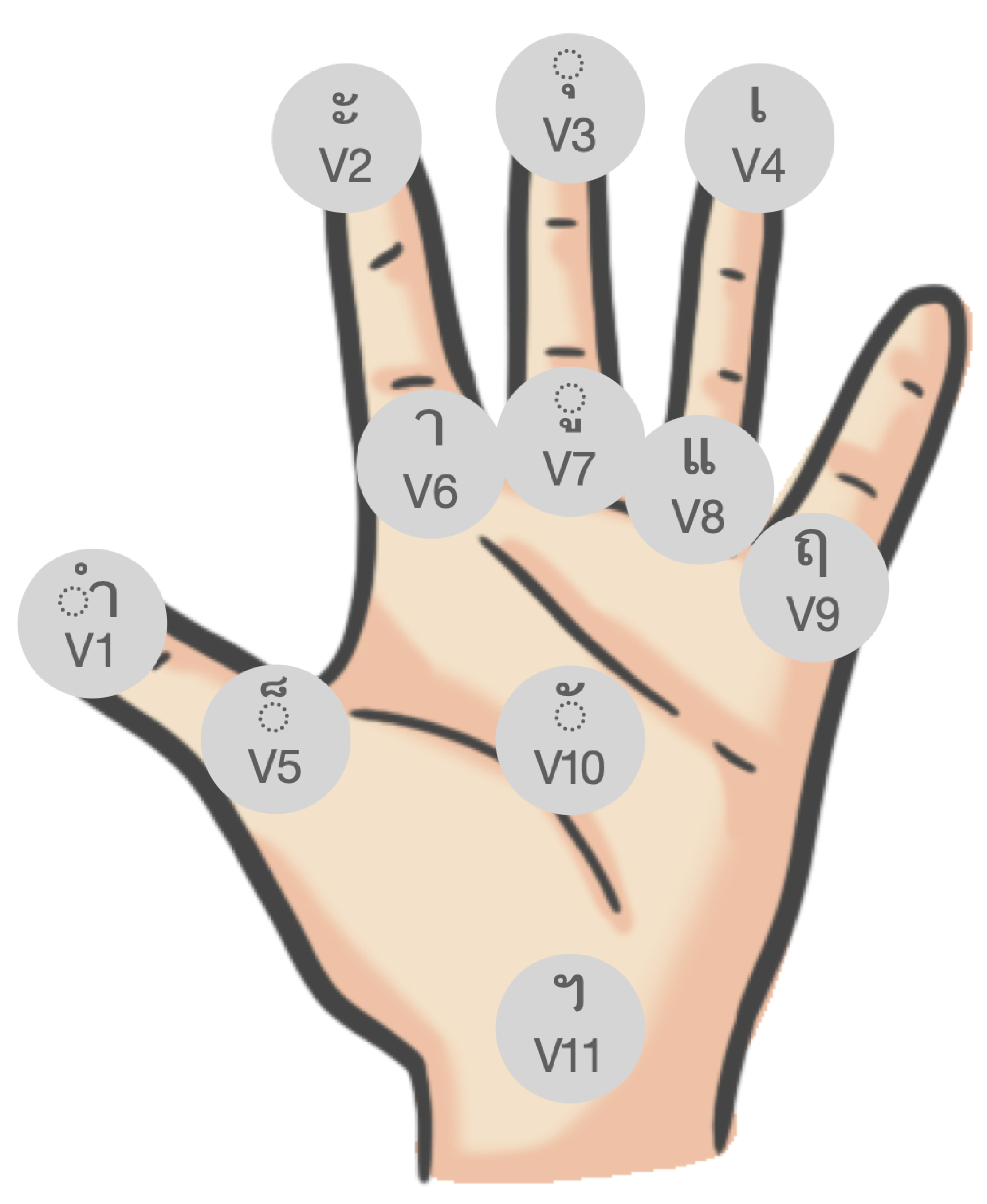}
		\caption{Designated areas of TFS signs in the static-point-on-hand scheme}
		\label{fig:TFS-keypoints}
	\end{figure}

A static-single-hand scheme signs a character with a single pose using only one hand.
	%
A dynamic-single-hand scheme signs a character with movement or multiple poses using only one hand.
	%
A static-point-on-hand scheme requires precise localization of TFS keypoints. 
This scheme signs a character by pointing an index finger to a specific position on another hand
without movement---the second hand is posing in an outstretching open-palm posture. 
Figure~\ref{fig:TFS-keypoints} illustrates the designated areas on a hand using in a static-point-on-hand scheme.
%
	%
A dynamic-point-on-hand scheme signs a character in a similar way 
to a static-point-on-hand scheme but with a movement. 

Note that
a signer can use either left or right hand for both static- and dynamic-single-hand schemes
as well as switching hands in both point-on-hand schemes.



TFS sign recognition has gained some research attention\cite{NK2021}.
Most previous works\cite{TFS_plain, TFS_vgg, TFS_yolo, NK2021}
rely on training recognition models from scratch.
However,  
MediaPipe Hands (MPH),
high precision hand-landmark localization
provided by Google\cite{MPH},
has
recently
been made publicly available.
MPH has been developed through multiview bootstrapping,
whose essential is using specialized equipment to facilitate training
keypoint detection in 3D space.
The advent of publicly available
hand-landmark localization
may allow a rapid development of a hand sign recognition system.

Our approach examines an approach built upon 
MPH architecture and its readily trained weights for TFS recognition. 
To further process the MPH hand landmarks to a TFS sign,
Artificial Neural Networks (ANNs)
are trained as a classifier
for static-single-hand scheme and a simplified version of dynamic single hand scheme (\textsection~\ref{sec: MPH-ANN}).
Relying on finger vectors and relative thresholding,
a rule-based method is developed 
to address a static-point-on-hand scheme (\textsection~\ref{sec: MPH-RB}).

Our study attempts to address three out of four TFS schemes
and examine MPH 
for being a feature extractor in a TFS recognition system
and for its own performance under presence of self occlusion.

	\section{Related Works}
	Thai Finger Spelling can be categorized into four different schemes
	as shown in Figure~\ref{fig:gesture-vs-scheme}.
	%
%
A static-single-hand scheme is commonly addressed through Convolutional Neural Network (CNN), either dedicated as a classifier using along with hand localization or framed it as an object detection homogeneously handling both localization and classification.
	Nakjai and Katanyukul\cite{TFS_plain, TFS_vgg} develop a CNN model to recognize TFS single-hand signing on plain black background. 
They handle the dynamic postures by simplifying movement into a few static postures. 
	%
	%
	Their pipeline begins with hand localization based on skin color before feeding the cropped hand image into CNN
	with fully connected layers to classify 25 classes of TFS. 
Their models have achieved 91.26\% and 97.59\% accuracies using their customized CNN and standard VGG-16\cite{vgg16}, respectively. 
	Later study\cite{TFS_yolo} addresses TFS on a complex background using YOLO\cite{yolo9000} object detection and achieves mAP of 82.06\%.
	
	Pariwat and Seresangtakul\cite{TFS_strokes} develop a CNN-based framework for static- and dynamic-single-hand in TFS. 
They categorize TFS signs into three sets---one stroke, two strokes and three strokes depending on its signing movement. 
Also capable of handling complex background, their framework begins with hand localization using SegNet\cite{segnet}.
	Then, they classify each single cropped hand image by AlexNet\cite{alexnet}.
Unlike earlier works\cite{TFS_plain, TFS_vgg, TFS_yolo}, Pariwat and Seresangtakul do not simplify the dynamic-single-hand scheme. 
Given the signing video, they use optical flow along with their customized rules
to address the dynamic-single-hand scheme
and have achieved 88.00\%, 85.42\% and 75.00\% accuracies for one-, two- and three-strokes, respectively.
	%

	Out of four TFS schemes, two single-hand schemes have been sufficiently studied. However, there is no previous work addressing  any TFS point-on-hand scheme.  
	Addressing point-on-hand schemes may require locating hand keypoints in high precision.
	%
InterHand\cite{interhand} and MediaPipe Hands (MPH\cite{MPH}) are among the notable off-the-shelf precise hand-keypoint detections.

InterHand\cite{interhand} is to predict 3D hand keypoints from an image.
The InterHand model has been trained with annotations of 42 hand keypoints (21 keypoints for each hand).
Its mechanism relies on Gaussian blob\cite{3d_confidence}
to generate a 3D confidence map of each keypoint.
InterHand has not been employed in any sign language recognition 
(at the time of this writing).

MPH\cite{MPH}---targeted for AR/VR applications---%
breaks down the problem of detecting hand keypoints into sub-problems of detecting hands, then detecting keypoints on every hand found.
This presumably leads to efficiency and high accuracy.
The approach	
starts from detect a hand, then detect keypoints. They use three strategies to detect a hand: 
	(1) detecting a palm, whose shape is more definite,
	rather than detecting a full hand with fingers, whose appearances
	are quite amorphous, to reduce the problem complexity; 
	(2) using encoder-decoder model to
	handle the small-context of a case when a hand size appears small in the image;
	%
	(3) reducing the penalty to mitigate big losses when a hand size appears large.
	%
	Given a cropped hand image, MPH predicts 21 keypoints using
	multiview bootstrapping (MB)\cite{Multiview_Bootstrapping} whose model has been prepared with images of multiple views acquired at the same time using 
	the multi-camera dome. 
	Using this MB, the model has learned to associate the target to images of different viewpoints.
	In addition, training image annotations of different viewpoints have been mutually corrected through triangulation. 
The model itself is topologically Convolutional Pose Machine (CPM\cite{posemachine}). 
	
	CPM predicts each keypoint as a confidence map. 
	%
	This architecture also improves the quality of a predicted confidence map using series of stage modules---each module is a small CNN.
%
	Stage modules refine their outputs---sets of confidence maps---in a successive manner such that confidence maps of the later stages achieve finer detail than ones of the earlier stages. 
	%
To mitigate an issue of missing keypoints,
the missing keypoints have been masked out in the training loss,
so that any prediction (or an absence of it) on a missing keypoint would be left un-penalized.

	MPH has been employed to address finger spelling recognition in many other sign languages with high accuracy\cite{asl_mph,  many_mph}.
	Shin et al.\cite{asl_mph} address American Finger Spelling (AFS) using MPH to extract 21 hand keypoints that are converted to distance and angle features. 
	The features then are classified into AFS signs using Support Vector Machine (SVM)\cite{svm} and light Gradient Boosting Machine (GBM)\cite{light_gbm}.
	Their systems are evaluated by accuracy on three datasets, i.e., 99.39\% on the Massy dataset\cite{data_asl_massy}, 87.60\% on the ASL Alphabet dataset\cite{data_asl_alphabet} and 98.45\% on Finger Spelling A dataset\cite{data_asl_finger_spelling_A}.
	%
	%
	Halder and Tayade\cite{many_mph} use MPH along with various classifiers to addresss American, Indian, Italian and Turkish Finger Spelling with up to 99\% accuracy.

Sign languages addressed above 
have only static signing scheme with no posture with self occlusion (hands obstructing each other). 
TFS has four schemes covering static and dynamic postures with and without self occlusion. 
Therefore, addressing complete TFS signing may requires various strategies than these previous works.

	The advent of a well-trained MPH can provide a quick jump-start for developing TFS sign recognition.
However, CPM---the technology underlying MB upon which MPH is built---was reported degradation when target objects are in close proximity.
	Since point-on-hand signing highly involves hand-hand interaction---close proximity, or sometimes even self occlusion, of both hands---, this may affect MPH performance as we employ it for the point-on-hand scheme. 
	Our study is also set out to explore this point.

From an architectural perspective,	
our study examines a higher level design combining
a well-established module with a task-specific component
for TFS recognition.
MPH is a relatively general module explored here
for its suitability on various TFS schemes.
From a task perspective,
our work attempts to address
the unexplored point-on-hand scheme,
as well as examines a new approach for well studied single-hand schemes.

	\section{MPH-based approach}
	
\begin{figure}[H]
		\centering
		\includegraphics[width=0.8\textwidth]{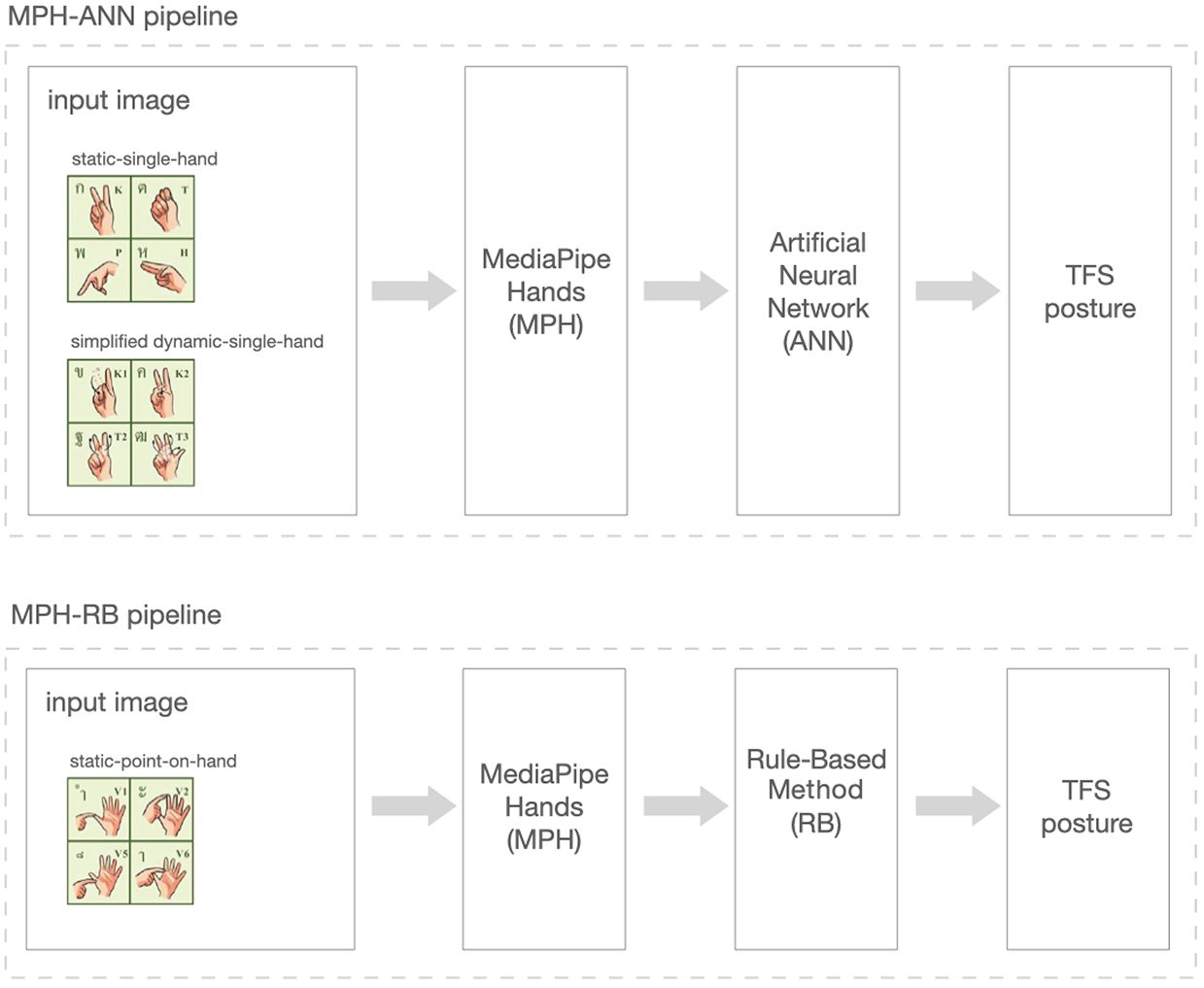}
		\caption{MPH-ANN pipeline (for 
		static- and dynamic-single-hand schemes)
		and
		MPH-RB pipeline (for a static-point-on-hand scheme).}
		\label{fig:pipelines}
	\end{figure}

	We examine MPH using in two pipelines (Figure~\ref{fig:pipelines})
to address three TFS schemes, i.e., a static-single-hand scheme, a dynamic-single-hand scheme and a static-point-on-hand scheme.
The first pipeline employs Artificial Neural Network (ANN) for both static- and dynamic-single-hand schemes. 
The second pipeline employs Rule-Based Method (RB) for a static-point-on-hand scheme.
	
	Both pipelines use hand landmarks from MPH as their input.
We have developed a switching rule
to decide which pipeline will be used.
The switching rule is to classify the landmarks into either
a single-hand case or a point-on-hand case. 
The single-hand case is responded by the MPH-ANN pipeline, 
while the point-on-hand case is taken by the MPH-RB pipeline. 
		
	\subsection{Switching rule}

The switching rule
checks a number of appearing hand, $H$, in the MPH output.
If $H$ is one, then it is the single-hand case
and the MPH-ANN pipeline is activated.
If $H$ is two, then
it is the point-on-hand case
and the MPH-RB pipeline is activated.

	
	\subsection{MPH-ANN pipeline}
	\label{sec: MPH-ANN}
	
Once activated, the MPH-ANN pipeline
passes the hand landmarks to an ANN model for predicting a TFS sign. 
The MPH-ANN pipeline is designed for a still image. 
Following previous works\cite{TFS_plain, NK2018},
we simplify a dynamic-single-hand scheme by
taking
a series of snapshots from signing video data.
This simplification allows an easier task and a homogeneous application of the pipeline to both static and dynamic schemes.
	
	
	
	
	
	%
	The MPH provides extracted image features---21 landmarks each of which is three dimensional.
	%
	%
We investigate two versions of the ANN: absolute- 
and relative-coordinate versions.   
Both versions use a three-fully-connected-layer network with 60, 40 and 30 hidden units, respectively. 
The output layer uses sigmoid as its activation function, 
the others use rectified linear unit (relu) with batchnorm.

The only difference between two versions 
is whether the hand landmarks feeding to the ANN are absolute or relative coordinates.
The absolute version takes a 63-dimensional vector%
---representing 21 absolute coordinates---directly obtained from MPH.

The relative version takes a 60-dimensional vector derived from 20 relative coordinates---using wrist landmark as a reference.
That is, denote a landmark vector $\mathbf{A} = [\mathbf{a}_1, \ldots,  \mathbf{a}_N]^T$,
where
$\mathbf{a}_i$ 
is a three-dimensional coordinate of the $i^{th}$ landmark
and $N$ is a number of landmarks (63 in our case).
The relative coordinates
$\mathbf{a'}_i = \mathbf{r} - \mathbf{a}_i$
for any $i^{th}$ landmark that is not a reference,
where $\mathbf{r}$ is the reference coordinate.
	
	
	
	\subsection{MPH-RB pipeline}
	\label{sec: MPH-RB}

Once activated, the MPH-RB pipeline deduces a TFS sign
from the hand landmarks
using a set of rules presented as pseudocode 
in shown Table~\ref{tab:pseudocode}.

\begin{table}[]
	\begin{tabular}{|l|}
		\hline
		\begin{minipage}{1\textwidth}
			
			\begin{verbatim}
				Choose hyper-parameter THRES
				Input M: image

				// 1. Get output from MPH
				// MPH returns a set of landmark vectors.
				// Each vector corresponds to each hand detected.
				SetA = MPH(M)
				
				// 2. Both hands are detected
				IF has_two_hand(SetA) 
				
				   // 3. Both hands are in static-point-on-hand scheme
				   IF is_PoH(SetA) 
				
				       Ap = get_pointing_hand(SetA)
				       Ao = get_open_palm_hand(SetA)
				       P = get_index_finger_tip(Ap)
				       // 4. Get the nearest landmark to the pointing tip
				       NL = get_nearest_landmark(P, Ao) 
				
				       // 5. Check proximity to the nearest landmark 
				       // is in threshold 
				       IF is_in_threshold(P, NL, THRES)
				
				           // 6. Predict the sign of nearest landmark
				           prediction = get_sign(NL)
				
				       ENDIF
				   ENDIF
			ENDIF
			\end{verbatim}
			
		\end{minipage}
		\\ \hline
	\end{tabular}
\caption{Rule-based method for MPH-RB pipeline}
\label{tab:pseudocode}
\end{table}
	 
(Step 1) The MPH-RB pipeline takes hand landmarks from MPH. 
(Step 2) It checks a number of detected hands.
It can proceed only if both hands are detected. 
(Step 3) It checks if hand poses are in the static-point-on-hand scheme. 
(Steps 4-6) It checks if the distance between a tip of the pointing finger tip and the nearest landmark is less than a specified threshold. 
If the distance falls within the threshold,
it identified a TFS sign by the nearest landmark.

Specifically,
given an image $\mathbf{M}$,
MPH provides a mapping $\mathbf{M} \mapsto \{\mathbf{A}_1, \ldots, \mathbf{A}_H\}$,
where $\mathbf{A}_i$ is a 63-hand-landmark vector
of the $i^{th}$ detected hand
and $H$ is a number of hands detected.
Note that MPH also provides handedness of every hand it detects,
but we omit other information unrelated to our rule-based method for brevity.

Function \verb|has_two_hand| (Step 2) can easily be achieved by checking $H$.
Function \verb|is_PoH| (Step 3) identifies a point-on-hand posture
when one hand is posing 
an outstretching open-palm posture
and another is pointing.

\begin{figure}[H]
\centering
\includegraphics[width=0.4\textwidth]{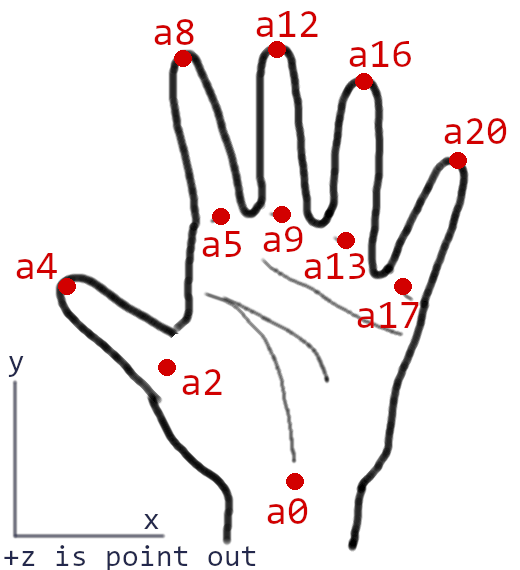}
\caption{Identifying an open-palm posture.}
\label{fig: PoH open palm}
\end{figure}

\paragraph{Rule 3.1 (open-palm condition)}%
To determine
an open-palm posture,
we conjecture that
an open-palm signing is posed upright and 
tips of the four fingers lie above their knuckles,
while the thumb tip lies opposite the wrist 
with the thumb knuckle lying in between (Figure~\ref{fig: PoH open palm}).
That is, identify a landmark vector $\mathbf{A}$ 
as an open palm
only when
\begin{eqnarray}
y(\mathbf{a}_8) &>& y(\mathbf{a}_5)
\nonumber, \\
y(\mathbf{a}_{12}) &>& y(\mathbf{a}_9)
\nonumber, \\
y(\mathbf{a}_{16}) &>& y(\mathbf{a}_{13})
\nonumber, \\
y(\mathbf{a}_{20}) &>& y(\mathbf{a}_{17})
\nonumber, \\
(x(\mathbf{a}_4) - x(\mathbf{a}_2)) \cdot (x(\mathbf{a}_0) - x(\mathbf{a}_2)) &<& 0
\label{eq: open palm condition}
\end{eqnarray}
where $y(\cdot)$ and $x(\cdot)$ are
y and x components of a coordinate, respectively.

\begin{figure}[H]
\centering
\includegraphics[width=0.4\textwidth]{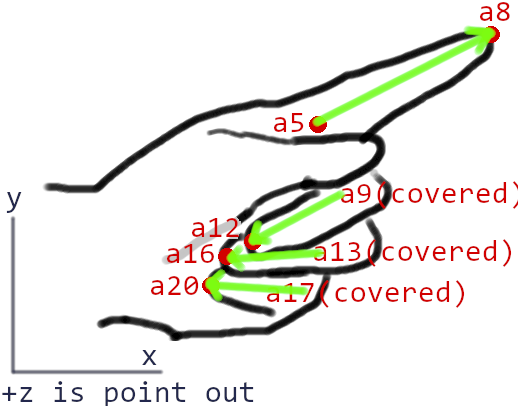}
\caption{Identifying a pointing posture.}
\label{fig: PoH pointing}
\end{figure}

\paragraph{Rule 3.2 (pointing-pose condition)}%
To determine pointing posture, we hypothesize that 
pointing is posed with index finger pointing out
while the other three fingers curling in
(Figure~\ref{fig: PoH pointing}).
Note that training with multiview bootstrapping 
MPH can provide hand-landmark coordinates, even ones that are occluded. 

Simplifying the matter into 2D,
z component is ignored.
Denote
an index-finger vector
as
$\mathbf{v}_8 = \mathbf{\hat{a}}_8 - \mathbf{\hat{a}}_5$,
where $\mathbf{\hat{a}}_i$ is an xy-coordinate of $\mathbf{a}$.
Similarly,
$\mathbf{v}_{12} = \mathbf{\hat{a}}_{12} - \mathbf{\hat{a}}_9$,
$\mathbf{v}_{16} = \mathbf{\hat{a}}_{16} - \mathbf{\hat{a}}_{13}$,
and
$\mathbf{v}_{20} = \mathbf{\hat{a}}_{20} - \mathbf{\hat{a}}_{17}$
represent
the other three fingers.
Identify a landmark vector $\mathbf{A}$ 
as a pointing pose
only when
\begin{eqnarray}
\mathbf{v}_8 \cdot \mathbf{v}_{12} &<& 0
\nonumber, \\
\mathbf{v}_8 \cdot \mathbf{v}_{16} &<& 0
\nonumber, \\
\mathbf{v}_8 \cdot \mathbf{v}_{20} &<& 0
\label{eq: pointing condition}.
\end{eqnarray}

Given Rules 3.1 and 3.2, 
a point-on-hand scheme can be determined,
as well as which hands are posing pointing or open palm.
Once a pointing hand is identified,
a coordinate of the pointing tip, $\mathbf{a}_p$, can easily be obtained.
So is its nearest landmark on the open-palm hand.
Euclidean distance is used for distance measure. 
Then, thresholding is performed to validate the signing.
For robustness,
we use a relative threshold,
i.e., 
a one-third of an average length of the four fingers (index, middle, ring and pinky) as the threshold.

\section{Datasets and Evaluations}
		
	The MPH-ANN pipeline is trained and evaluated using 
	a single-hand dataset: 2518 training and 1801 testing images at 1280 x 720 pixels.
	The single-hand dataset composes of thirty pose classes from static- and simplified dynamic-single-hand schemes. 
	In order to account for result variation, we perform bootstrapping 10 repeats for each version---absolute and relative versions (\textsection~\ref{sec: MPH-ANN}).
	We use accuracy to evaluate MPH-ANN pipeline and support it with significance tests.

	The MPH-RB pipeline is a rule-based method. Thus, it does not require training.
	We evaluate this pipeline with a point-on-hand dataset: 1488 images at 720 x 720 pixels. 
The point-on-hand dataset is composed of eleven classes representing poses in a static-point-on-hand scheme.
The accuracy is also used to evaluate MPH-RB pipeline.
	


	To provide a benchmark, another approach based on
classification is prepared and tested on TFS point-on-hand-sign recognition.
The classification approach is adapted from 
a TFS VGG-based classification\cite{TFS_vgg},
but configured to target 11 point-on-hand signs.
The point-on-hand classifier is examined on
two datasets.
The first dataset has 1650 images (1254 for training and 396 for testing) of 220 x 220 resolution.
All images are modified to have their background plain black.
This first dataset will represent an easier task.
The second dataset is derived from the first one 
by replacing the plain black background with randomly chosen natural  images.

	%

	\section{Results}
\begin{figure}[H]
		\centering
		\includegraphics[width=0.6\textwidth]{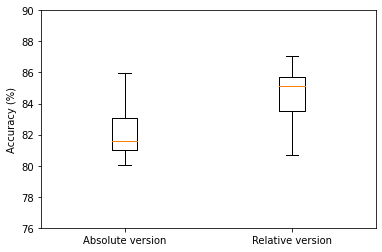}
		\caption{Boxplots of accuracies of both absolute and relative versions on single-hand schemes}
		\label{fig:single-hand-result}
	\end{figure}

	The MPH-ANN approach has effectively addressed single-hand schemes with accuracies of 82.12\% and 84.57\% using absolute and relative versions, respectively.
	Figure \ref{fig:single-hand-result} shows accuracies of the MPH-ANN pipeline on single-hand schemes in boxplots.
	The t-test confirms that the relative version significantly outperforms the absolute version with p-value 0.0057---normality of both versions were justified by the Shapiro–Wilk test\cite{shapiro_and_wilk} at 5\% significance level.

	%
	
\begin{figure}[H]
		\centering
		\includegraphics[width=0.5\textwidth]{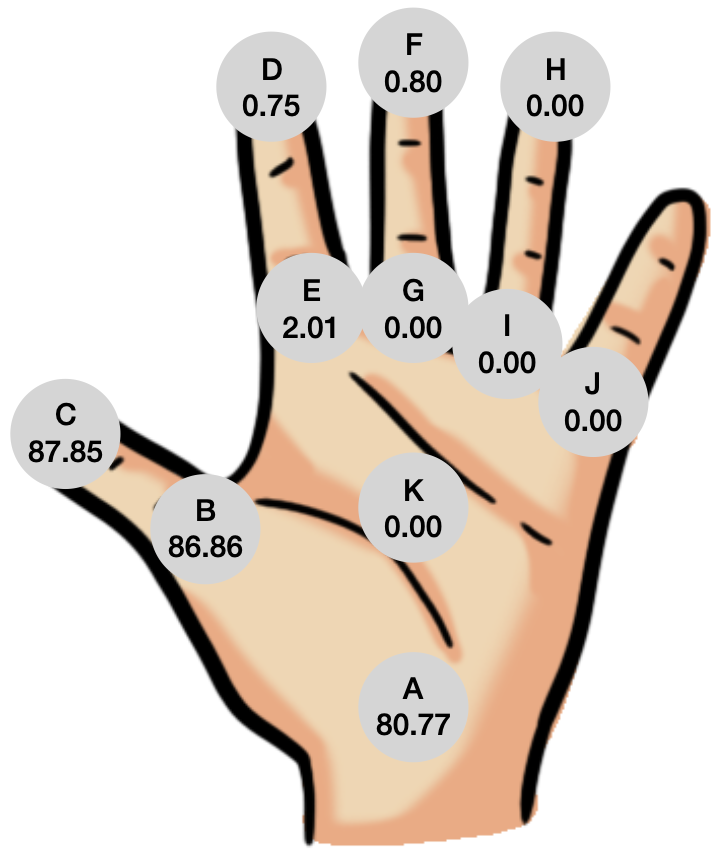} 
		\caption{Recognition results as true-positive percentages on a static-point-on-hand scheme.}
		\label{fig:poh_result}
	\end{figure}

	The MPH-RB approach fails to address the point-on-hand scheme, as reflected by its low accuracy of 23.66\%. The true positive percentage (TP) by keypoint is reported in Figure \ref{fig:poh_result}.
	On the near edge, we notice that keypoints A, B and C were well addressed with TPs of 80.77\%, 86.86\% and 87.85\%.
The system was not able to handle
the interior keypoints (D, E and F),
while completely failed on the furthest keypoints (G to K).

\begin{figure}[H]
		\centering
		\includegraphics[width=0.6\textwidth]{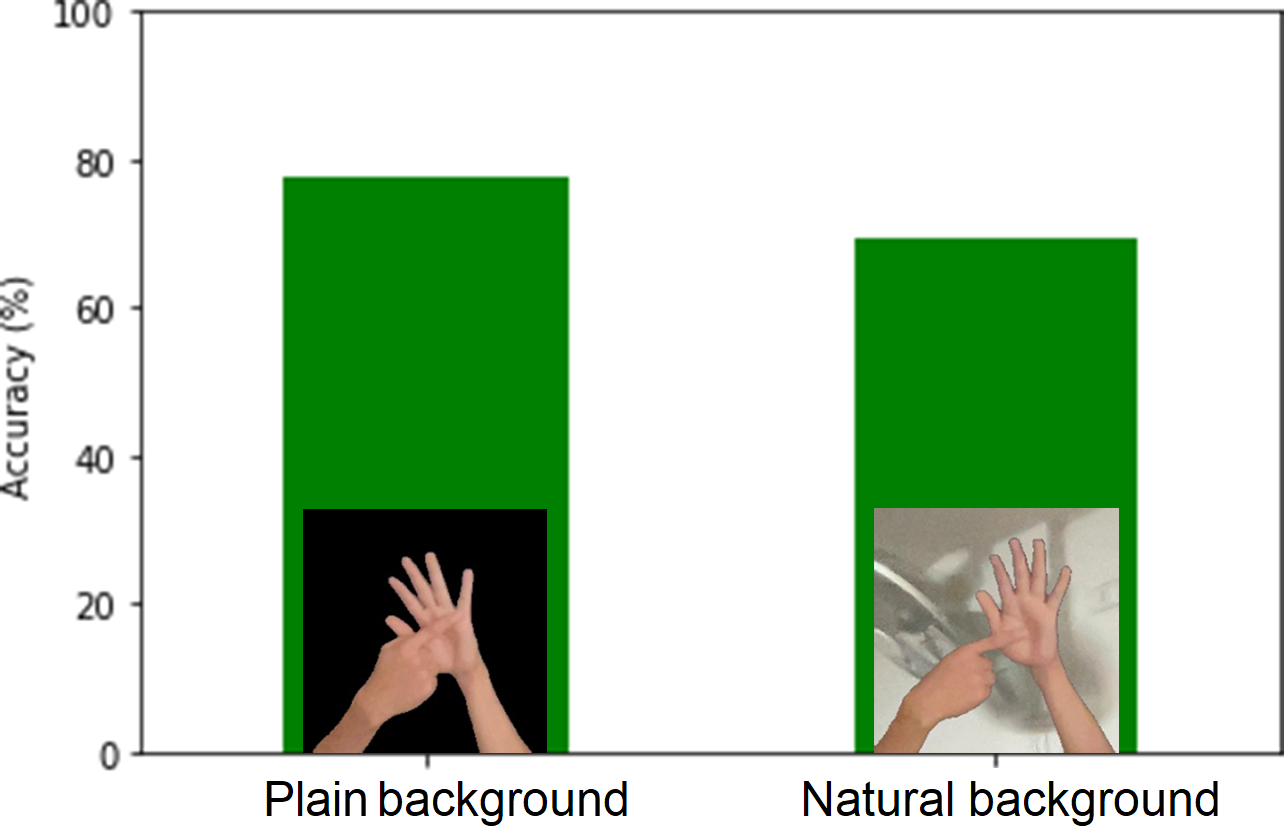}
		\caption{The test results of a classification approach on  point-on-hand sign recognition.}
		\label{fig:PoH-classification-no-bg}
	\end{figure}

To put MPH-RB performance in perspective,
the point-on-hand classifier 
has been tested and shown  
77.78\% and 69.19\% of accuracies on 
plain-background and natural background datasets, respectively (Figure~\ref{fig:PoH-classification-no-bg}).
Note that this point-on-hand classifier
has been trained and tested using different datasets than the one used for MPH-RB.

Despite the entire MPH-RB pipeline failure,
our quick post-inspection reveals
effectiveness of 
the rule-based method 
in determining an open-palm or pointing posture
when MPH gives dependable keypoints.
Therefore, 
we suspected that MPH may fail in one of the conditions appear when signing a point-on-hand scheme.
The further investigation of MPH is discussed in \textsection\ref{cautions}.

	\section{Discussion}

Apparently as shown in Figure~\ref{fig:single-hand-result}, static- and simplified dynamic-single-hand schemes
are well addressed using MPH through MPH-ANN pipeline.
The accuracies are well over 80\% using either absolute or relative version.
The relative-coordinate version gives about 2.45\% edge over the absolute-coordinate version.

On the other hand, application of MPH through MPH-RB pipeline clearly fails to address the static-point-on-hand scheme as shown by
the average accuracy of 23.66\% (conferring to 77.78\% and 69.19\% obtained from a classifier on similar tasks).
The error analysis (Figure~\ref{fig:poh_result}) reveals
a striking pattern.
The pipeline works fine on keypoints (A to C) near an edge that is close to another hand, which is signing a pointing gesture.
However, it works dramatically poorer for the interior keypoints
and completely fails on keypoints on the further edge.

We speculate hand-hand interaction, particularly self occlusion, as a possible underlying factor.
Pointing to the near-edge keypoints causes
little self occlusion,
while pointing to other keypoints 
causes significant self occlusion,
as the pointing hand needs to move in and occlude a part of the open-palm hand.
This assumption is consistent with the result that the further away the location of a keypoint is from the pointing hand,
the worse accuracy it turns out.
Section~\ref{cautions} discusses our 
further investigation on this issue.

Regarding performance of a point-on-hand classifier,
contrary to our initial belief
a conventional classification approach seems to do pretty well.
It can reach 77.78\% and 69.19\% accuracies
on similar but presumably easier tasks.
Note that this was achieved without much fine tuning or any additional mechanism or training trick.
It may disclose that TFS point-on-hand sign recognition might not be as highly challenging as we initially thought.

\section{Further Investigation} 
\label{cautions}

Our preliminary experiments showed that
self occlusion caused MPH serious deterioration in hand keypoint detection.
In addition, we saw incorrect prediction of handedness from MPH in some cases.
Both self occlusion and handedness issues are further investigated in \textsection\ref{sec: self-occlusion}
and \textsection\ref{sec: handedness}.

\subsection{MediaPipe Hands on Hand-Hand Interaction}
\label{sec: self-occlusion}

MPH is particularly examined on hand-hand interaction using a dedicated dataset  of 150 images at resolution 1280 x 720.
The images show postures posed by a signer 
with various degrees of self occlusion. 
The images are equally distributed into five groups each of 30 images:
images showing a single hand
and images showing two hand with 
0\%, 20\%, 50\%, and 80\% of self occlusion.
The images showing a single hand serve as a baseline case.

\begin{figure}[H]
		\centering
		\includegraphics[width=0.68\textwidth]{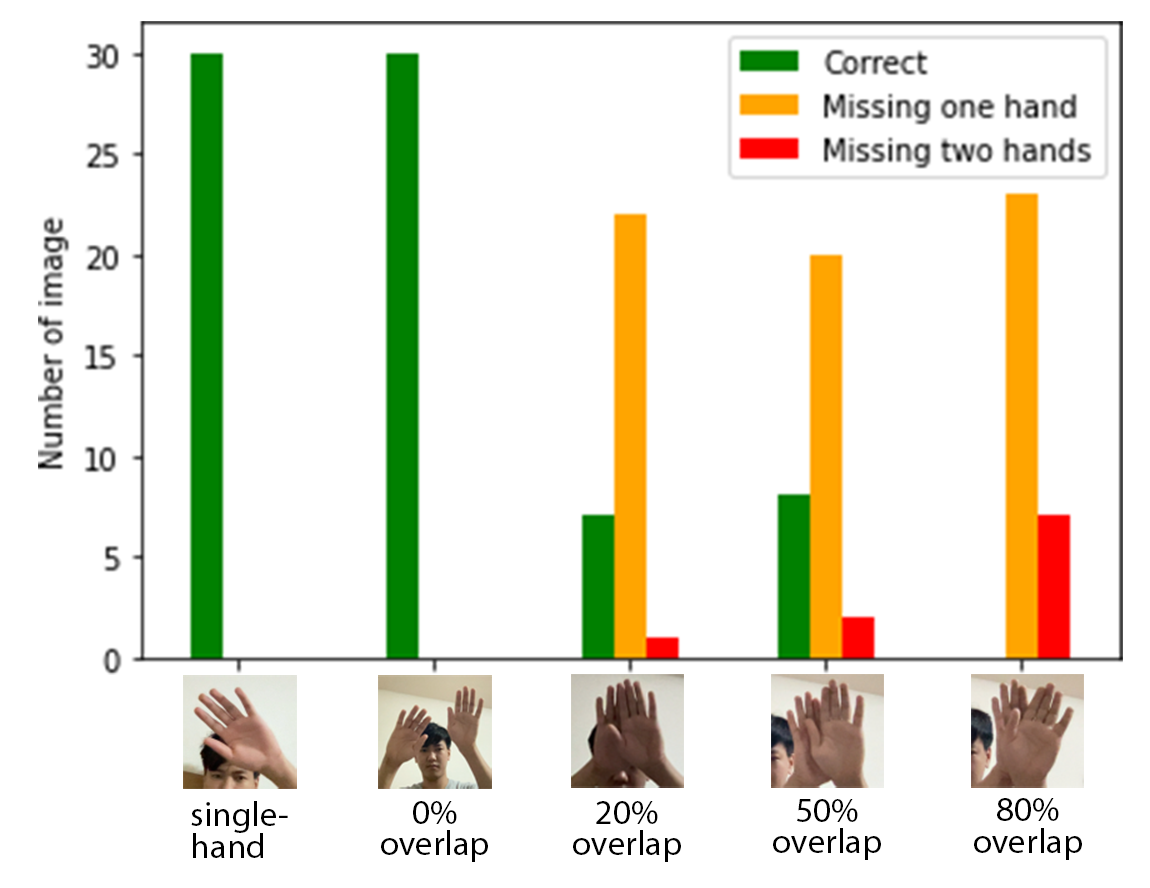}
		\caption{MPH on various degrees of occlusion.}
		\label{fig:result-MPH-overlap}
	\end{figure}

	The results are shown on Figure~\ref{fig:result-MPH-overlap}. 
MPH can correctly detect a hand or hands as well as keypoints 
when there is only one hand present (single hand case, the leftmost on the figure)
and 
when there are two hands but no occlusion (0\% occlusion case).
With only a small degree of occlusion, MPH performance seems to be hugely deteriorated,
as an evidence of a lower number of images that MPH gets the detection correctly (shown as in green),
a higher number of images that MPH fails even to detect a hand (shown in yellow) or both of them (shown in red).

This clearly supports our conjecture that MPH has a difficult time 
with occlusion.
Although occlusion generally is a challenging issue,
but this investigation shows that MPH effectiveness drops dramatically even with 20\% occlusion.

This weakness on self occlusion may be explained through MPH mechanism.
MPH mechanism starts from detecting hands before processing each hand for its corresponding keypoints.
This mechanism allows MPH to be very efficient and suited for a real-time application for which its design targets.
However, this design may make self occlusion
seriously inimical to MPH, 
especially when the occluding covers features of the occluded that MPH relies on.
Self occlusion makes a shape of the occluded appear less definite
and may directly challenge the very fundamental assumption of MPH that detecting palm is easier than detecting an entire hand because palm shape is more definite.

\subsection{MediaPipe Hands on Handedness}
\label{sec: handedness}

We investigate MPH on handedness using 
a dataset consisting of 240 360x360-resolution images of a single signer.
It is to examine MPH handedness recognition on eight posture categories, e.g.,
showing a front view of a right hand (FR), 
showing a front view of a left hand and
a back view of a right hand (FLBR).
	%
Each posture category is represented by 30 images.

\begin{figure}[H]
		\centering
		\includegraphics[width=0.5\textwidth]{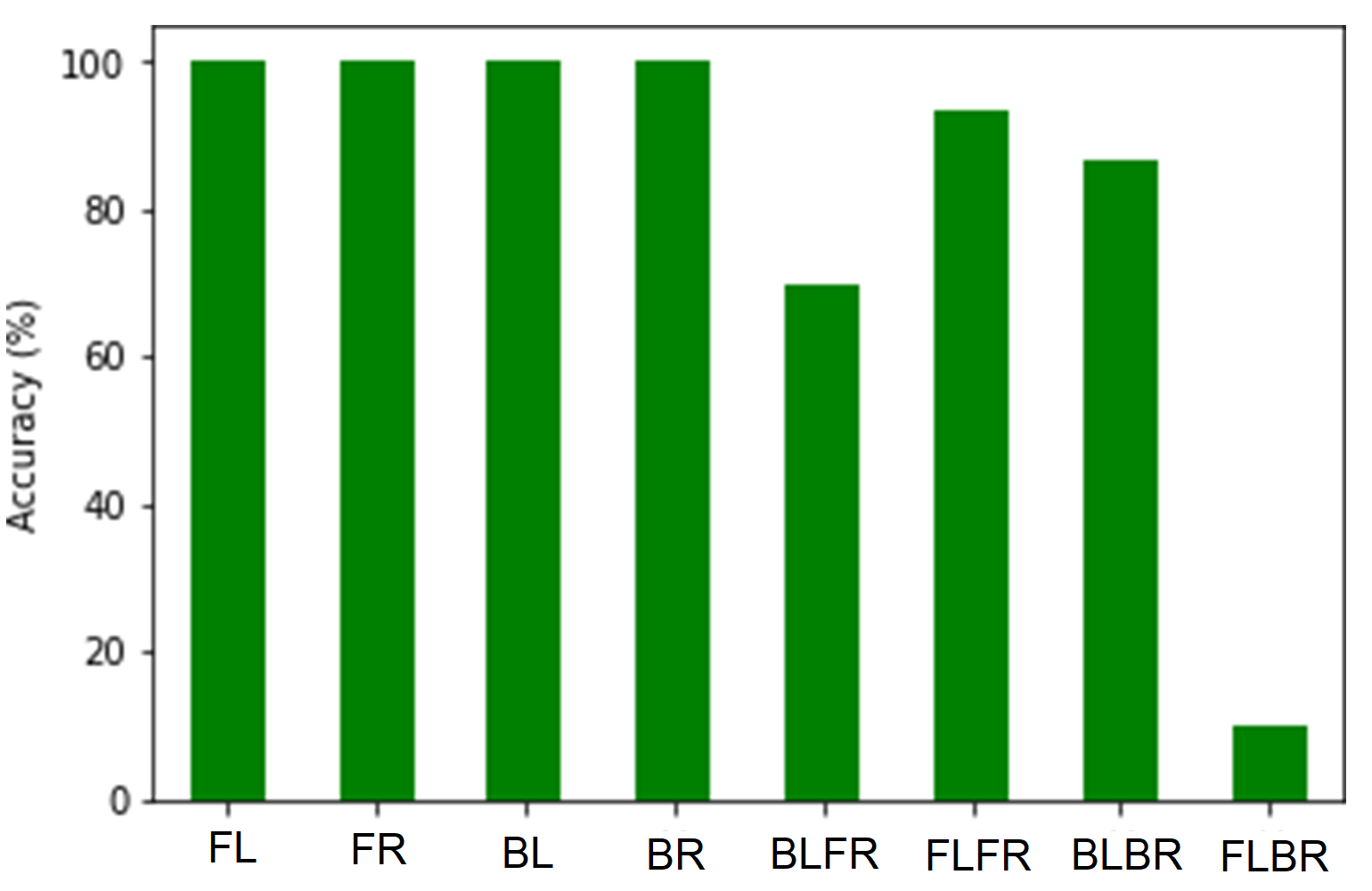}
		\caption{Performance of MediaPipe Hands on Handedness Prediction}
		\label{fig:result-MPH-left-right}
	\end{figure}

Figure~\ref{fig:result-MPH-left-right}
shows accuracies of MPH prediction on handedness.
	
Apparently, MPH is most effective for single hand cases (FL, FR, BL, and BR):
no matter the hand is posed a front view or a back view, 
MPH can predict handedness accurately.

Surprisingly, when both hands are posing,
MPH accuracies drop variably.
It seems like 
MPH is doing fine when both hand pose the same view:
93.33\% for FLFR
and
86.67\% for BLBR.
But, when both hand pose different views,
it seems to confuse MPH:
70.00\% for BLFR
and
10.00\% for FLBR.
	
	
	\section{Conclusion and Prospective Studies}

	We have investigated applications of MediaPipe Hands (MPH) to Thai Finger Spelling (TFS) recognition.
	Specifically, MPH is used as the hand-landmarks-feature extractor
	in our proposed pipelines.
The TFS static- and dynamic-single-hand	schemes are satisfactorily addressed through a pipeline (84.57\% accuracy).
The pipeline uses a fully-connected Artificial Neural Network (ANN) to process hand keypoints obtained from MPH.
We also found that a variation using relation distances among keypoints, rather than using keypoint coordinates directly, lead to better performance.

Regarding a point-on-hand scheme,
an application of MPH fails to address TFS static-point-on-hand scheme.
However, a classification approach has shown to be able to achieve 77.78\% and 69.19\% accuracies on plain and complex backgrounds, respectively.

The failure of an application of MPH to point-on-hand sign recognition
have been further investigated 
and the factor underlying failure has been attributed to a shortcoming of MPH under
self occlusion.
In addition to occlusion, 
handedness when both hands are present
seems to be an issue for MPH.
The handedness issue can get worse if both hands are posing different views, i.e., one showing a front view while another showing a back view.
Given our findings,
MPH can provide hand keypoints very fast
and detection is of high quality when only hand involves.
However,
its application should be cautious 
when there is a chance of both hands showing up in an image frame.
MPH functionality may fail totally when self occlusion is present.

For prospective studies,
beside MPH, another 
well-established 
hand-keypoint detector
is
InterHand\cite{interhand}.
While MPH architectural design is detecting hands then processing each one independently,
InterHand is intrinsically designed to account for both hands.
It could address the issue that MPH has with a point-on-hand scheme.

In addition, despite that
our MPH-based pipeline on a point-on-hand scheme fails, 
our proposed method in the pipeline
seems to function well given dependable keypoints.
Therefore, exploring an application of InterHand
along with our method based on finger vectors seems highly prospective.

Also, an extension to cover TFS dynamic schemes---%
i.e. dynamic-single-hand scheme (without simplification) and dynamic-point-on-hand scheme---%
would be worth a dedicated study.



	\bibliography{sk02}



\end{document}